\documentclass[runningheads]{llncss}
\usepackage{graphicx}
\usepackage{amsmath}
\usepackage{amssymb}
\usepackage{comment}
\usepackage{subcaption}
\usepackage[font={footnotesize,it},labelfont=bf]{caption}
\usepackage{accents}
\usepackage{algorithm,algorithmic}
\usepackage{booktabs}
\usepackage{bm}
\usepackage{enumitem}
\usepackage{multirow}
\usepackage[table,xcdraw]{xcolor}

\newcommand{\R}{\mathbb{R}}

\newcommand{\E}{\mathbb{E}}

\usepackage{color, colortbl}
\definecolor{LightCyan}{rgb}{0.88,1,1}
\definecolor{Gray}{gray}{0.9}

\usepackage[normalem]{ulem}

\begin{document}

\title{AI Liability Insurance With an Example in AI-Powered E-diagnosis System}

\titlerunning{AI LiabilityInsurance}

\author{
Yunfei Ge\inst{1} \and
Quanyan Zhu\inst{1}}
\authorrunning{Ge and Zhu}
\institute{New York University, Brooklyn, NY 11201, USA \\
\email{\{yg2047,qz494\}@nyu.edu}}
\maketitle              

\begin{abstract}

Artificial Intelligence (AI) has received an increasing amount of attention in multiple areas. The uncertainties and risks in AI-powered systems have created reluctance in their wild adoption. As an economic solution to compensate for potential damages, AI liability insurance is a promising market to enhance the integration of AI into daily life. In this work, we use an AI-powered E-diagnosis system as an example to study AI liability insurance. We provide a quantitative risk assessment model with evidence-based numerical analysis. We discuss the insurability criteria for AI technologies and suggest necessary adjustments to accommodate the features of AI products. We show that AI liability insurance can act as a regulatory mechanism to incentivize compliant behaviors and serve as a certificate of high-quality AI systems. Furthermore, we suggest premium adjustment to reflect the dynamic evolution of the inherent uncertainty in AI. Moral hazard problems are discussed and suggestions for AI liability insurance are provided.  

\keywords{AI Insurance, Cyber Insurance, Security Economics, \and Artificial Intelligence}
\end{abstract}

\section{Introduction}

Artificial Intelligence (AI) is an emerging technology that has been utilized in various areas, such as autonomous vehicles, healthcare, security, and many others. As AI continues to develop, it has the potential to transform traditional industries and improve outcomes in many ways. However, the risks and uncertainties associated with AI have restricted the wide adaptation of AI technology \cite{kelly2019key}. For example, there have been instances where the AI-powered supercomputer recommended erroneous treatment recommendations \cite{ross2018ibm}. 
Such incidents highlight the need for appropriate measures to be taken to minimize risks and ensure the safety and reliability of AI systems. This is where AI liability insurance comes in. An appropriate market for AI-enabled technologies must include insurance as a tool to incentivize regulated behaviors and hedge against the risks of AI-inflicted damage \cite{lior2022insuring}. AI liability insurance can protect individuals, businesses, and society at large from potential harm caused by AI systems, and can offer financial compensation for damages incurred. Furthermore, AI liability insurance can encourage responsible behavior and practices among AI developers and manufacturers. By requiring them to obtain liability insurance, it incentivizes them to take appropriate measures to minimize risks and ensure the safety and reliability of their AI systems. This can help promote the responsible and ethical development of AI technology, which in turn can lead to increased trust and confidence in these systems.

Despite the necessity of AI liability insurance, there is, however, limited research in insuring AI systems as yet. The main concern is the associated risks. As an emerging technology, the insurance company and market currently lack information about what damage AI systems can cause. Besides, the decision-making process of an AI product cannot be evaluated while the decision is being made \cite{desai2017trust}. The inherent uncertainties in the ``black-box'' or ``grey-box'' decisions create unpredictability of AI technologies and make it difficult to design the insurance plan. Moreover, risk quantification is hard as the damages of AI systems could be widespread and cross-functional.

Existing insurance plans cannot serve the purpose of AI liability insurance. Different from general \textit{liability insurance}, AI liability insurance should consider the inherent uncertainty during the design of the insurance plan. It is important to note that this uncertainty will abate over time as people have more knowledge about AI. The insurance plan should adjust the premium flexibly based on the knowledge of the uncertainty. \textit{Cyber insurance} is another framework that could involve AI liability. However, most cyber insurance insures against the outcome of an exogenous attack while AI liability insurance insures against the inherent liability of the product. In addition, as we will discuss AI-powered E-diagnosis systems in this work, we would like to emphasize the difference from \textit{medical malpractice insurance}. Medical malpractice insurance is limited to individuals, but AI liability can propagate to a population of users and patients. As for now, there are few studies on AI liability insurance in classification systems. We would like to extend the research and focus on the AI E-diagnosis system in this work.

To this end, we use an AI-powered E-diagnosis system as an example to study AI liability insurance. A quantitative risk assessment model is provided with numerical analysis. Both diagnosis performance risks and the inherent uncertainty in AI are considered. We use practical data and machine learning models to support the quantification. 
We define the insurability requirements for the AI E-diagnosis systems and discuss the dominating conditions for insurability. Furthermore, we suggest dynamic premium adjustment to reflect the dynamic features of the inherent uncertainty in AI. Moral hazard problems are discussed and suggestions for AI liability insurance are provided.

We suggest building upon existing insurance frameworks with necessary adjustments to accommodate the unique features of AI products. AI liability insurance can act as a regulatory mechanism to incentivize compliant behaviors of AI entities. It can also serve a certification role for AI products, enabling the adoption of those technologies that can be demonstrated to have limited or quantifiable risk. By considering dynamic premium adjustment, we take into account the inherent uncertainty in AI and ensure the insurance company is profitable. AI Insurance provides a way to compensate for the potential risks and losses within AI technology. This encourages early adopters of innovation to pay the premium to hedge their risky bets and activities. In general, the establishment of AI liability insurance will promote AI adoption and reduce AI-inflicted risks.

The rest of the paper is organized as follows. The related works are provided in Section~\ref{sec:related}. Section~\ref{sec:model} provides an overview of the AI liability insurance model and risk assessment. We present the optimal insurance plan and its influencing factors in Section~\ref{sec:analysis}. AI insurability analyses with numerical examples are provided in Section~\ref{sec:result}. We finally conclude the paper in Section~\ref{sec:conclusion}.

\section{Related Work}
\label{sec:related}

There are two major types of risks in an AI-driven technology: intentional cyber risks and unintentional liability risks. Cyber risks refer to the risks that come from an exogenous attack, for instance, data breach \cite{talesh2018data}, ransomware attack \cite{zhao2021combating}, open source software vulnerability attack, etc. This type of risk can be covered by first-party or third-party cyber insurance \cite{marotta2017cyber}, which mainly covers the related loss in the cyber domain after attacks. On the other hand, in liability risks, the AI system fails due to its inherent unpredictability or bad quality without any adversarial tampering. The liability risks of AI services can lead to performance degradation and losses beyond the cyber domain. For instance, the fatal crash caused by Uber's autonomous vehicle in 2020 \cite{cellan2020uber} is an example of an AI system causing physical damage to a third party. This type of loss is not likely to be covered by cyber insurance as they are not directly related to data breaches or abuse. Hence, liability insurance related to AI technology should be developed to fill the gap.

As for now, there are different forms of insurance that include AI liability. In the area of autonomous vehicles, Tesla Motor launched its in-house insurance program in 2019 to cover the liability of its products by the company itself \cite{tesla}. 
As for AI algorithms, some of the cyber insurances included algorithm liability loss as part of cyber loss and provide insurance as an element under cyber insurance \cite{romanosky2019content}. For AI-enabled products, some insurance plans directly utilize existing liability insurance frameworks without considering the unique features of AI technologies \cite{josephdamage}. None of the approaches considers the inherent AI uncertainty and unpredictability in the insurance plan and provides solutions tailored for AI-powered systems.

Existing studies on insurance in the AI realm mainly focus on the social impact and potential influences that AI insurance could bring.
Lior in \cite{lior2022insuring} discussed the advantages and disadvantages of utilizing insurance in the AI market. Stern et al. addressed how AI liability insurance can facilitate the adoption of AI technologies in healthcare and reduce predictable risks \cite{stern2022ai}.  Tournas et al. proposed that AI insurance can serve as a soft law that governs AI technologies \cite{tournas2021ai}.
There is a lack of work providing quantitative suggestions for AI liability insurance plans. Our work differentiates from existing work as we dive into the details of AI classification algorithms and discuss the insurability and suggestions for AI-driven technologies.

\section{Insurance Model}
\label{sec:model}

In this work, we advocate third-party liability insurance for AI technologies. Liability insurance will cover the potential risks due to the inherent uncertainty and errors within the AI system. We use the AI-powered E-diagnosis system as an example to analyze influencing factors in detection systems and discuss the optimal policies given different situations.

\begin{figure}[!t]
    \centering
    \includegraphics[width=0.8\linewidth]{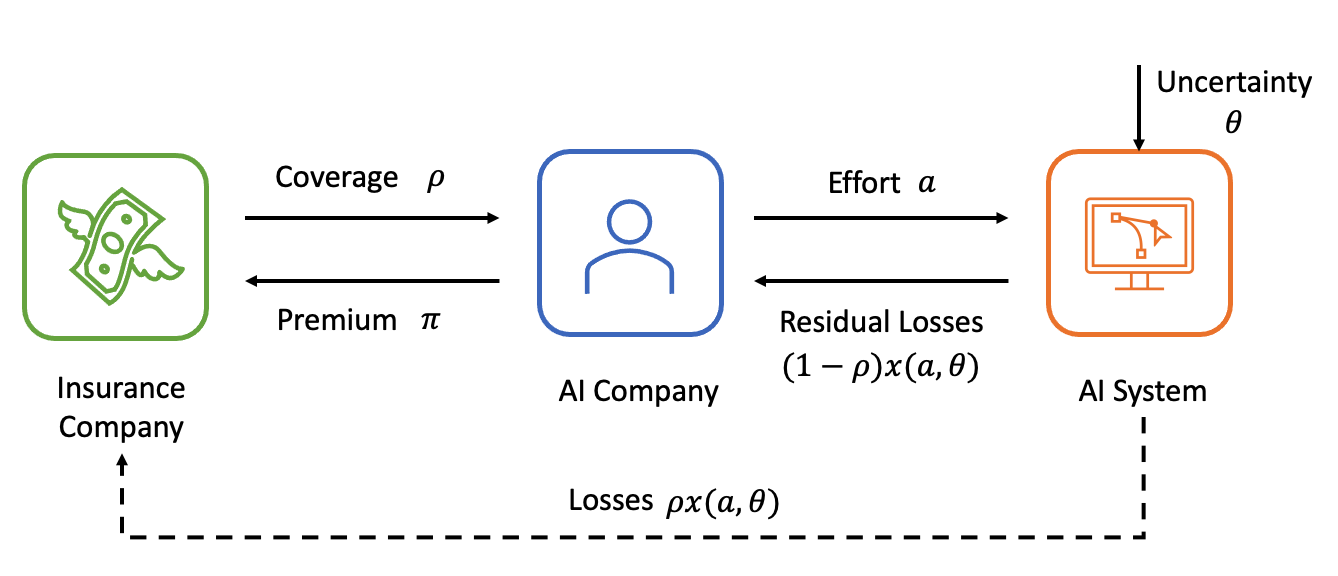}
    \caption{Basic AI liability Insurance Model. The uncertainty of the AI system $\theta$ is a publicly known random variable that cannot be controlled through the AI company's effort $a$.}
    \label{fig:plan}
\end{figure}

\subsection{Basic Insurance Model}

E-diagnosis is a type of medical diagnostic system that automatically generates test results based on samples. It is built upon well-vetted computational algorithms. AI offers a principled approach to developing such algorithms for the analysis of medical data. The performance of the system depends on the company's investment in system design and algorithm development. 
Suppose the AI company chooses an amount of effort $a\in \mathcal{A}$ to improve the e-diagnosis system. The cost of the effort is given by a cost function $c(\cdot):\mathcal{A}\mapsto \R_+$
The effort a is \textit{hidden action} that cannot be observed by the insurer. Instead, the insurer can see the \textit{outcome} $x = X(a,\theta)$. The outcome $x$ can be viewed as the loss of the AI system. 

The random variable $\theta\sim\Sigma$ represents the inherent uncertainty in the AI system as the ``black-box'' decision-making process is unforeseeable. This uncertainty represents the lack of information about AI decisions at the current stage. By including this value in the insurance design, we can provide an upper bound on risk estimation and ensure the profitability of the insurance company. 
It should be noted that this lack of knowledge will abate over time as AI technologies are more widely studied.
Due to the uncertainty in the AI decision-making process, we assume the risk $x$ admits to a parameterized distribution function $F(x|a)$ suppressing $\theta$. A natural assumption is that $F_a(x,a)>0$, for $\forall x\in\mathcal{X}$, i.e., an increase in the effort $a$ shifts some probability weight from higher to lower values of $x$.

Let $\Omega$ be the set of observable and contractable events. In this model, the insurance company knows the distribution of outcome $F(x|\cdot)$ and the distribution of uncertainty $\Sigma$, but cannot observe the effort of the AI company $a$ and the realization of $\theta\sim\Sigma$.
In a linear insurance contract, the insurance company offers a premium $\pi\in\R_+$ and a coverage level $\rho\in [0,1]$ to the agent. Thus, the insurance contract is a mapping $\Omega \mapsto \R_+\times[0,1]$. The basic insurance model is illustrated in Figure~\ref{fig:plan}.

Consider the insured agent (AI company) admits to a utility function $U(\cdot)$ 
and the insurance company admits to a risk-neutral utility function $V(\cdot)$, i.e., $V(x)=x$. Suppose the initial wealth of the insurance company is $w_I\in \mathbb{R}_+$ and the initial wealth of the agent is $w_A\in\mathbb{R}_+$. 
 $\underline{U}$ is the minimum utility threshold that the agent can bear. The optimal insurance contract is given by 
\begin{equation}
\begin{aligned}
 \max_{\pi,\rho,a} \quad & \int V\left(w_I + \pi -\rho x(a) \right) d F(x|a) & \\
\textrm{s.t.} \quad & \int U\left(w_A - c(a)-\pi-(1-\rho)x(a)\right) d F(x|a) \geq \underline{U} & \textrm{(IR)}\\
& a\in\arg\max_{a'\in\mathcal{A}} \, \int U\left(w_A - c(a)-\pi-(1-\rho)x(a)\right)d F(x|a) & \textrm{(IC)}
\end{aligned}
\label{eq:original}
\end{equation}
The first constraint is the individual rationality (IR) constraint to ensure the agent would participate in the game. The second constraint is the incentive compatibility (IC) constraint that outputs the utility-wise optimal effort for the agent.

\subsection{Risk assessment of AI-powered E-diagnosis system}

In general, the risk of an AI-powered system can be expressed as the risk aggregation of the system distribution
\begin{align}
    x(a,\theta) = D_{AI}\times (l_{AI}(a) + l_{\theta})
    \label{eq:risk}
\end{align}
where $D_{AI}$ is the user population size of the AI system.  
We divide the risks of a single AI system into two parts. $l_{AI}(a)$ is the loss of a single operational error given the company's effort $a$. This value is controllable as the company can put more effort into algorithm development to reduce this loss. Another risk is the inherent uncertainty in the AI black-box decision that cannot be controlled denoted by  $l_{\theta}$.
In this work, we assume the probability distribution $l_{AI}(a)\sim f(x|a)$ and $l_\theta\sim f(\theta)$ are independent. 

In the following section, we will use the AI-powered E-diagnosis system as an example and discuss each component in \eqref{eq:risk}. We use logistic regression as the AI classification model and discuss the relationship between the company's effort and system performance. We use the Covid-19 E-diagnosis as an application and estimate the potential loss based on real-world evidence.
    
\begin{figure}[!t]
    \centering
    \includegraphics[width=0.9\linewidth]{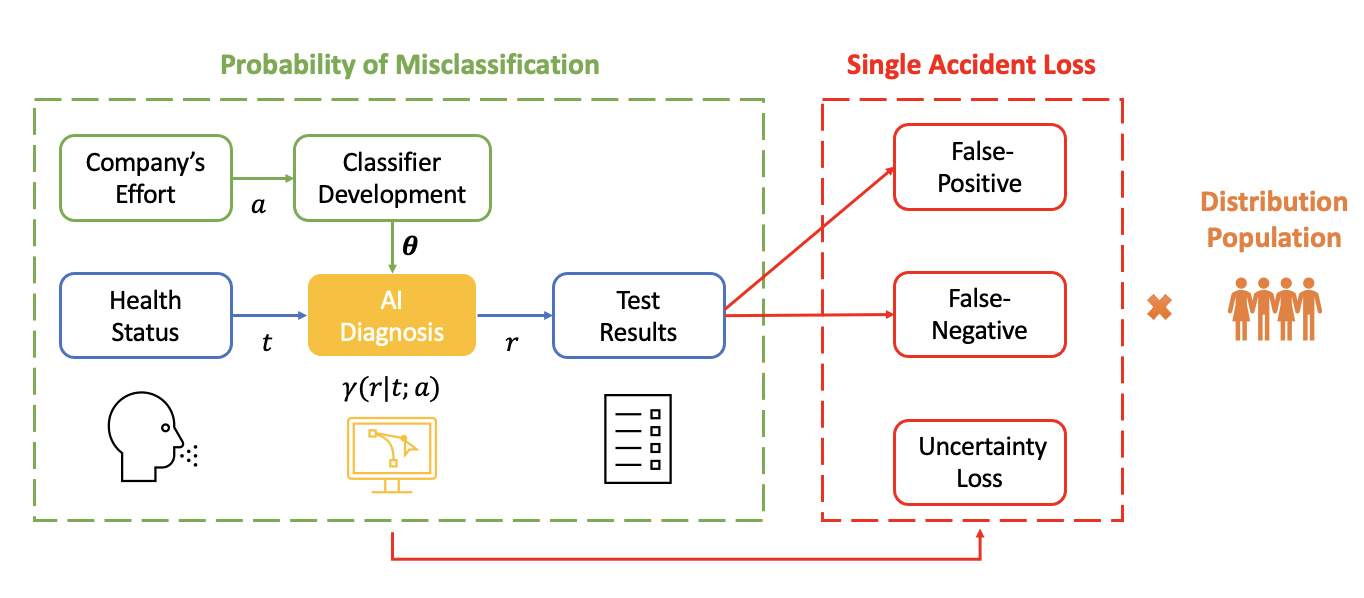}
    \caption{Risk assessment of AI-powered E-diagnosis system. The risks occur when the system outputs erroneous test results or the system operates unexpectedly due to AI uncertainty.}
    \label{fig:riskassess}
\end{figure}

\subsubsection{Population size $D_{AI}$}

The user population size $D_{AI}$v depends on how many AI products have been distributed. This is the key difference between medical malpractice and AI insurance. Malpractice insurance is limited to individual cases but AI liability can propagate to hundreds of patients or more.  
Automated detection accelerates the diagnosis process and greatly saves time for the patients. In the example of CT imagining diagnosis, AI-enabled diagnosis only takes $0.744$ minutes on average while conventional doctor diagnosis spends the average time of $3.623$ minutes \cite{zhang2021value}. Although AI-powered diagnosis improves testing efficiency, conversely, the malfunction of the AI system can create a larger impact compared to the malpractice of a single doctor. 
Consider a doctor who makes false diagnoses with $D_{doc}$ patients within a period. Suppose the AI diagnosis runs $k>1$ times faster than a conventional doctor. With only one AI diagnosis machine, the diagnosis error can influence $K\cdot D_{doc}$ patients within the same period. If the company has distributed $n$ AI diagnosis machine, the user population size influenced by AI misinformation has the following relationship compared to the population size influenced by a single doctor:
\begin{align}
    D_{AI} = k\cdot n\cdot D_{doc}.
\end{align}
Depending on the efficiency of the AI diagnosis and the product distribution, a malfunction in AI diagnosis could have a greater impact on society. Thus, it is important to establish an insurance mechanism to support the market.

\subsubsection{Single accident  loss $l_{AI}(a)$}

\noindent \textbf{Classification Performance (loss probability):}

We assume the AI-powered E-diagnosis system aims to perform a binary classification task based on the received test sample. Let $t\in \mathcal{T}=\{0,1\}$ be the true health state of the patient, where $t=0$ means that the patient is uninfected, and $t=1$ means that the patient is infected.  The AI-powered e-diagnosis system is a classifier that assigns a diagnosis to a given patient based on the observed characteristics. We use $s_t\in \mathcal{S}$ to represent the test sample from the patient with health status $t$. The system receives the test sample and outputs the test result $r\in \mathcal{R}=\{0,1\}$, where $r=0$ means ``diagnosed negative,'' and $r=1$ means ``diagnosed positive''. 

Given an unknown distribution $\mathcal{D}$ on the labeled sample space $ \mathcal{S}\times \mathcal{T}$. Let $\mathcal{Z}=\{z_1,z_2,\dots,z_N\}$ denote $N$ samples $z_i = (s_i,t_i)$ drawn i.i.d. from $\mathcal{D}$.  Let 
$\gamma:\mathcal{D}\mapsto\mathcal{T}$ be the target classification function. 
Let $l(t,r=\gamma(s))$ be the loss function, where $t$ is the true health status and $r=\gamma(s)$ is the prediction from the system.
Given the sample data $\mathcal{Z}$ and the classifier $\gamma$, the empirical risk is 
\begin{align}
    \E_{(s,t)\sim \mathcal{Z}}\left[l(t,\gamma(s))\right]=\frac{1}{N}\sum_{i=1}^{N} l(t_i,\gamma(s_i)).
\end{align}
The goal of training is to find a classifier such that the empirical risk is minimized. 
\begin{align}
    \gamma^* = \arg\min_{\gamma} \E_{(s,t)\sim \mathcal{Z}}\left[l(t,\gamma(s))\right].
\end{align}

In this work, we consider Logistic Regression (LR) classification model and the Stochastic Gradient Decent (SGD) method for learning purposes. In LR, the loss function is the non-convex log loss.  The performance of the developed classifier $\gamma$ can be evaluated through the generalization error
\begin{align}
   I(\gamma) = |\E_{(s,t)\sim \mathcal{Z}}\left[l(t,\gamma)\right]-\E_{(s,t)\sim \mathcal{D}}\left[l(t,\gamma)\right]|,
\end{align}
where $\mathcal{Z}$ is the empirical sample distribution and $\mathcal{D}$ is the target population distribution. It is shown in \cite{zhang2022stability} that with a non-convex loss function, the generalization error is tightly bounded by $\Theta (T^e/N^{1+e})$. $T$ is the number of training iterations, $e>0$ comes from the SGD step size $\alpha^t = e/(\beta t)$ with smoothness parameter $\beta$, and $N$ is the total number of samples. More training samples help reduce the generalization error and improve classification performance. 

AI companies can put effort into the classifier development stage to improve classification performance. In general, the effort can include data collection, model selection, algorithm design, etc. In this work, we assume the effort of the AI company is related to the data collection stage. We assume the number of training data $N$ is an increasing function with respect to the company's effort: $N = h(a)$, where $h:\mathcal{A}\mapsto \R_+$ and $h'>0$. Gathering more training samples (i.i.d. from target distribution) helps reduce the risk of misclassification, but will induce a cost to collect the data.

The logistic regression function outputs the probability that the sample $s_t$ is from an infected patient $t=1$, i.e., $\Pr[t=1|s_t]$. To generate the final classification decision, the system still needs to determine the threshold $\tau\in[0,1]$ such that
\begin{equation*}
    \gamma(s_t)=
    \begin{cases}
    r=1 & \text{if }\Pr[t=1|s_t]\geq \tau,\\
    r=0 & \text{otherwise,}
    \end{cases} \qquad \forall s\in\mathcal{S}.
\end{equation*}
Based on the choice of $\tau$, there are two types of errors in the system:
\begin{align*}
    &\text{Type I (False Alarm): } \,p_{f} = \Pr[r=1|t=0],\\
    &\text{Type II (Miss): }  \qquad\quad p_{m} = \Pr[r=0|t=1].
\end{align*}
It is worth noting that as the threshold $\tau$ decreases, the system is more capable of correctly classifying positive patients, false alarm rate $p_f$ will also increase. There is a fundamental trade-off between true-positive rate $p_t = 1-p_{m}\in[0,1]$ and false-positive rate $p_f$ depending on the classification design. 

A well-known method to describe the classification performance is the Receiver Operating Characteristic curve (ROC curve), which is a plot that describes the relationship between true-positive rate $p_t\in[0,1]$ and false-positive rate $p_f\in[0,1]$ in the square $[0,1]\times[0,1]$. The area under the ROC curve (AUC) is a measure of investigation capability that varies from $0.5$ to $1$. When AUC equals $0.5$, the developed classifier has no separation capability. Ideally, a perfect classifier will produce an AUC equal to $1$. From the previous discussion, the training process and the company's effort determine the AUC value of the classification model, while the choice of classification threshold $\tau$ determines the error rates under the model. 

\begin{figure}[t!]
\centering     
\subfloat[ROC]{\includegraphics[width=0.49\textwidth]{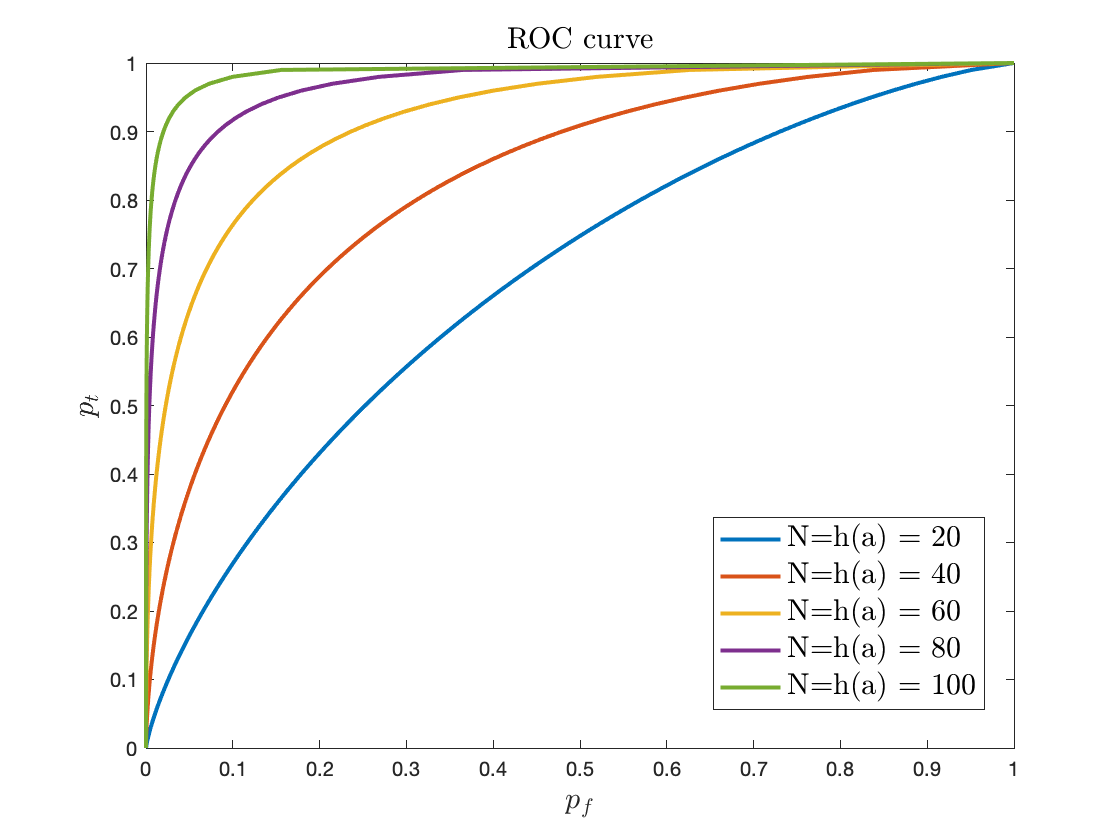}}
\subfloat[AUC]{\includegraphics[width=0.49\textwidth]{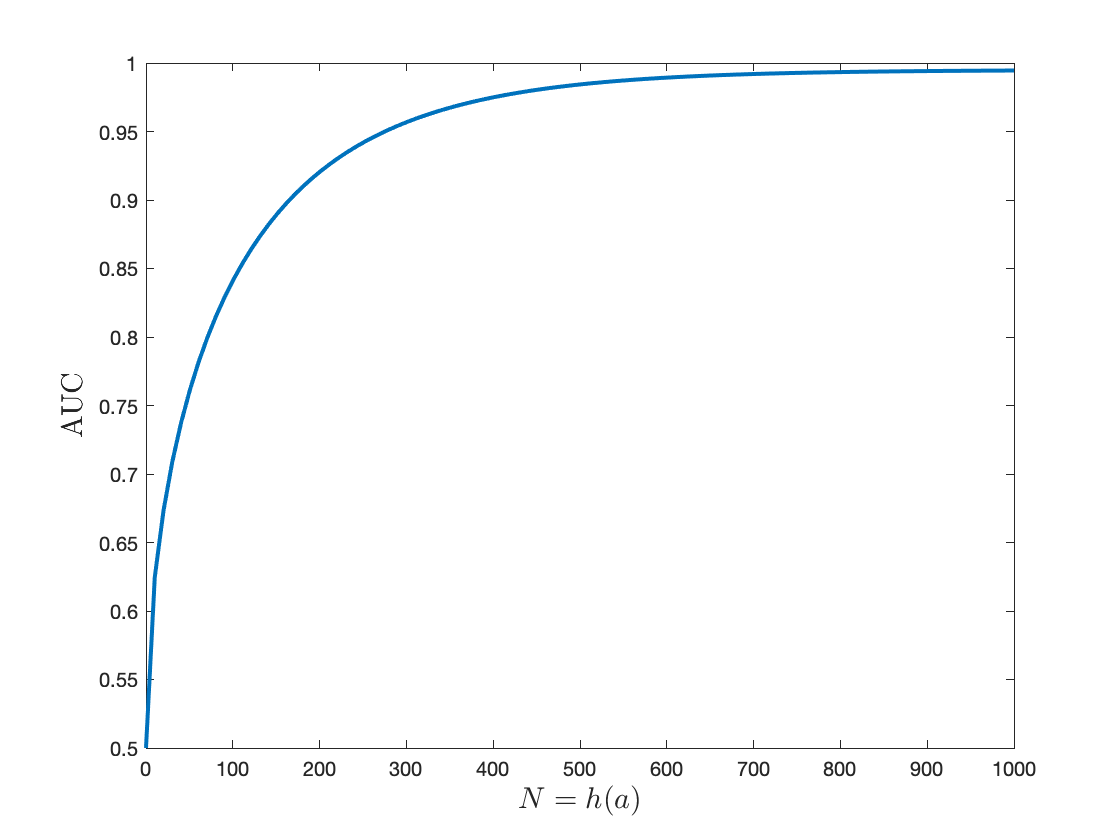}}
\caption{Classification performances with different sample size $N=h(a)$. More samples would improve the classification performance and increase the AUC value.}
\label{fig:roc}
\end{figure}

The optimal choice of classification method and threshold is beyond the discussion of the paper. 
For illustration purposes, we use the following expression to describe the relationship between the false-positive rate $p_f$ and the true-positive rate $p_t = 1-p_{m}$ \cite{levy2008binary}:
\begin{align}
    p_f(a) = Q(d(a) - Q^{-1}(1-p_t)), \qquad  \text{where } d(a) = D\cdot h(a).
\end{align}
$Q(x)$ is the Gaussian $Q$ function, $D\in\R^+$ is a constant related to the distance between the distribution $D(s_{t_1},\cdot)$ and $D(s_{t_2},\cdot)$, and $N = h(a)$ is the total number of samples related to the company's effort.  An illustration of the relationship between the company's effort and the classification performance is in Figure~\ref{fig:roc}.
If the system pick a fair classification threshold $\tau$ corresponding to the point which has the smallest distance to $(0,1)$ on ROC \cite{unal2017defining}, the two types of errors can be expressed as
\begin{align}
    p_f(a) = p_m(a) = Q\left( D\cdot h(a)\right).
\label{eq:pfpm}
\end{align}

\begin{remark}[Fundamental limitation on AI E-diagnoses performance]
    There exist fundamental limitations on the performance of AI-powered E-diagnosis systems. 
    Except for situations where the distribution $D(s_{t_1},\cdot)$ and $D(s_{t_2},\cdot)$ are completely separable or the number of samples $N=h(a)$ goes to infinity, the performance of the classification will be restricted within a feasible region. In general, it is impossible to find a perfect classifier with $100\%$ accuracy. Besides, the AI system performance is also limited by available computation. In the past decade, these constraints have relaxed along with specialized hardware (e.g. GPUs). However, because of the computational needs and available data scale so rapidly, there remain limitations on the computational power to find the optimal classifier. Mathematically, the population risk of the developed optimal classifier $\hat{\gamma}^*$ does not necessarily equals to zero:
    \begin{align}
        \Bar{L}(\hat{\gamma}^*) = \E_{(s,t)\sim \mathcal{D}}\left[l(t,\hat{\gamma}^*)\right]\neq 0
    \end{align}
    where $N$ is the number of data samples and $\mathcal{D}$ is the true distribution on the labeled sample space.
\end{remark}

\vskip 3mm
\noindent\textbf{AI-inflicted Damage Estimation (loss value):}\\
Two types of losses could occur due to misclassification:
\begin{itemize}
    \item \textbf{Case 1: }The patient is health $t=0$ but diagnosed positive $r=1$.\\
    This type of user would suffer a loss from unnecessary isolation and work restriction, as they can only work from home (WFH) instead of going to the office.  Suppose the original productivity per day is $W$, the required quarantine time is $T$, and the WFH productivity loss ratio is $\alpha \in (0,1]$, the false-positive losses on average would be 
    \begin{align}
        l_{FP}  =  \alpha W T .
    \end{align}
    According to \cite{gibbs2021work}, productivity fell by about $20\%$ working from home compared to the productivity before the COVID-19 pandemic. We can set the WFH productivity loss ratio $\alpha=0.2$ to estimate the productivity loss due to false-positive test results. We use the average daily salary in the U.S. to estimate the original productivity, thus $W = \$ 27\times 8 \textrm{hours} = \$ 216$. The required quarantine time is $T=5$ days.

    \vskip 2mm
    \item \textbf{Case 2: }The patient is infected $t=1$ but diagnosed negative $r=0$. \\
    Risks to a patient of a false-negative result include: delayed or lack of supportive treatment, unexpected loss in productivity due to COVID infection, and an increased risk of spread of COVID-19 within the community. Suppose the loss due to delayed treatment is $M$, the COVID productivity loss ratio is $\beta\in(0,1]$ and the cost of spreading is $C$, the false-negative losses on average would be 
    \begin{align}
        l_{FN} = M + \beta W T + C.
    \end{align}
    The delayed treatment could result in worsening medical conditions of the patient. We use the median cost for COVID-19 diagnostic test $M = \$148$ as an estimate of the delayed treatment loss \cite{mprice}.  COVID infection could create persistent symptoms (notably breathlessness and excessive fatigue) and limitations in reported physical ability \cite{arnold2021patient}. We can set the COVID productivity loss ratio $\beta=0.4$ to estimate the unexpected productivity loss due to the false-negative test result. To model the cost of spreading to other people, we consider the basic reproduction number, $R_0$, which is defined as the expected number of secondary cases produced by a single (typical) infection in a completely susceptible population. Research suggests that the Omicron variant has an average basic reproduction number of $R_0 = 9.5$ \cite{liu2022effective}. The infected agent could unconsciously infect $R_0$ colleges and influence their productivity. Thus, we assume 
    \begin{align*}
        C = R_0\cdot ( \beta W T).
    \end{align*}
\end{itemize}
Note that the false negative loss should be larger than the false positive loss, $l_{FN}>l_{FP}$, as the potential loss of wrongly assigned the infected patient into diagnosed negative would be larger that the opposite case.  In such a design, the model could choose a classification threshold $\tau$ that is more conservative to false negative results. 

Given the classification performance and loss values, we can write the single accident loss as a random variable
\begin{align}
    l_{AI}(a) = \begin{cases}
        l_{FP} & w.p.\quad p_f(a) \\
        l_{FN} & w.p.\quad p_m(a) \\
        0 & \text{otherwise}.
    \end{cases}
\end{align}

\subsubsection{Loss of uncertainty $l_\theta$}
The inherent uncertainty of the AI model derive could be originated from multiple sources \cite{klas2018uncertainty}.  Here are three major categories of AI uncertainty:
\begin{itemize}
    \item \textbf{Model uncertainty:} Uncertainty related to model fit is caused by the fact that AI techniques provide empirical models that are only an approximation of the real (functional) relationship between the model input and its outcome. The accuracy of this approximation is limited due to the limited number of model parameters, input variables considered, available data points in training, etc.
    
    \item \textbf{Sample precision:} In practice, all data collected (e.g., based on sensors or human input) is limited in its accuracy and potentially affected by various kinds of quality issues. The actual level of uncertainty in the outcome of an AI model is thus affected by the quality of the training data. This type of uncertainty can be reduced with balanced and adequate sampling.

    \item\textbf{Context Restriction:} AI models are built for and tested in a specific context. If the model is applied outside the scope for which it was developed, it can bring uncertainty to the final result. For instance, the AI E-diagnosis system developed based on the COVID Delta variant may perform worse with the Omicron variant in classification accuracy.

    \item\textbf{Computational Limitation:} As for now, most AI systems are strongly reliant on computing power. The AI system performance will be limited by available computation at the time. Although the constraints have been relaxed with the development of hardware, it is still a major problem for most AI systems.
\end{itemize}

Some of the inherent uncertainties are beyond the control of the agent. 

These uncertainties will be mitigated over time as people conduct more research about AI technology and obtain more data from implementation. By considering the risk associated with the inherent uncertainty in the insurance plan, we can provide an upper bound on risk estimation and ensure the profitability of the insurance company.

Without loss of generality, we assume the uncertainty is a random variable with Gaussian distribution with zero mean $\theta\sim \mathcal{N}(0,\sigma^2)$. Further, we assume the uncertainty always brings positive loss, thus the uncertainty loss.  We use the quadratic function to represent the loss of uncertainty. Given $\theta\sim \mathcal{N}(0,\sigma^2)$, we have
\begin{align}
    l_\theta = v\cdot \theta^2  \quad \Rightarrow \quad  \Bar{l}_{\theta} = \E_{\theta}[l_\theta] = v\sigma^2,
\end{align}
where $v\in\R_+$ is the amount of loss value caused by the uncertainty and $\sigma^2$ is the variance of the uncertainty.

\begin{remark}[Evolution of uncertainty $\theta$]
    As discussed before, the inherent uncertainty of the AI system arises from multiple aspects during the system design. To establish AI insurance, the market should have the ability to calculate the risks. 
    The past experiences and analysis set a baseline for initial uncertainty assessment in the AI context. This information should be updated with the development of AI systems.
    The uncertainty would be reduced or partially overcome when new information incorporating the enhanced capabilities of AI-based products has been collected and analyzed. Besides, the development of computational AI techniques will also reduce the unpredictability of AI systems and provide a more stabilized product with less uncertainty. 
    In our model, we can assume that the variance of $\theta\sim\mathcal{N}(0,\sigma^2)$ is decreasing with respect to time.
    \begin{align}
        \sigma^2(t) = \sigma^2_0 \exp(-m t) \quad (t>0),
        \label{eq:theta}
    \end{align}
    where $\sigma^2_0$ is the initial uncertainty of the AI system and $m$ is the learning rate of AI technologies. This formulation allows us to discuss the premium adjustment of AI insurance later.
\end{remark}

\section{Optimal Insurance Plan}
\label{sec:analysis}
\subsection{Full Information Benchmark}

Consider the following exponential utility function for the insurance participants:
\begin{equation}
    u(x)=\begin{cases}
    1-e^{-\epsilon x} & \epsilon \neq 0,\\
    x & \epsilon =0,
    \end{cases}
\end{equation}  
where $x$ is the wealth and $\epsilon \in \mathbb{R}$ is a constant that represents the degree of risk preference of the agent. $\epsilon>0$ represents risk averse; $\epsilon=0$ means risk neutral; $\epsilon<0$ is risk seeking.

When the effort of the company is observable and verifiable, there is no asymmetric information.  The optimal insurance contract is given by

\begin{equation}
\begin{aligned}
 \max_{\pi,\rho,a} \quad & \int V\left(w_I + \pi -\rho x(a) \right) d F(x|a) \\
\textrm{s.t.} \quad & \int U\left(w_A - c(a)-\pi-(1-\rho)x\right) d F(x|a) \geq \underline{U} 
\end{aligned}
\label{eq:firstbest}
\end{equation}
where $F(x|a)$ is the parametric probability measure of the loss suppressing $\theta$. 

From the previous discussion, the average loss is 
\begin{align}
    \E[x(a)] &= \int x d F(x|a)\\
    &= D_{AI} \left[ (p_f(a) l_{FP}+ p_m(a) l_{FN}) + v\int \theta^2  f(\theta) d\theta \right]\\
    &= D_{AI} \left[(p_f(a) l_{FP}+ p_m(a) l_{FN}) + v\sigma^2\right].
\end{align}
where $\sigma^2$ is the uncertainty variance. The first term in the bracket is the expected loss due to misclassification and the second term is the expected loss due to AI uncertainty. More effort the company put into system development and less uncertainty in AI technology will help reduce the expected risks of the AI-powered E-diagnosis system.

Let $\underline{U}=\E[U(w_A-c(a)-x(a))]$. Consider a risk natural insurer and a risk-averse agent with $\epsilon>0$.
Under full information ($a$ is known publicly), we can obtain the following theorem:
\begin{theorem} \label{theorem:full}
The insurance contract is established between the insurer and the agent with effort $a$ if the premium $\pi\in\mathbb{R}^+$ and the coverage level $\rho\in(0,1]$ satisfy
\begin{align}
    \rho \mathbb{E}[x(a)] \leq \pi \leq  \frac{1}{\epsilon}\log \mathbb{E}[e^{{\epsilon}\rho x(a)}]. 
\end{align}
The proof can be found in Appendix~\ref{ap:pro}. 
\end{theorem}

After determine the acceptable insurance plan $(\pi^*,\rho^*)$, the insurer prefer the first-best action $a^*$ such that the expected loss will be minimized:
\begin{align}
    a^* \in \arg\max_{a\in\mathcal{A}} \int V\left(w_I + \pi^* -\rho^* x(a) \right) d F(x|a) 
\end{align}

\subsection{Hidden action}
By solving equation \eqref{eq:firstbest}, we obtain the first-best insurance plan $(\pi^*,\rho^*)$ and the first-best effort $a^*$. 
After the announcement of the insurance plan, the agent will take rational action $a^\diamond$ such that:
\begin{align}
    a^\diamond \in\arg\max_{a'\in\mathcal{A}} \, \int U\left(w_A - c(a)-\pi^*-(1-\rho^*)x(a)\right) dF(x|a).
    \label{eq:mh}
\end{align}
This would be the optimal effort of the agent given the insurance plan  $(\pi^*,\rho^*)$. If we consider a risk-neutral agent, finding $a^\diamond$ is equivalent to solving
\begin{align}
     \min_{a\in \mathcal{A}} \quad c(a)+(1-\rho^*)D_{AI}\left[ p_f(a)l_{FP}+p_m(a)l_{FN}\right].
     \label{eq:eqopt}
\end{align}

We assume the effort is non-negative $a\geq0$ the cost function is a strictly convex function as $c'>0$ and $c''>0$. This indicates that the effectiveness of the effort in improvement decrease with the increase of the effort $a$. Further, we assume the two types of error rates are equal and satisfy $p_f(a)=p_m(a)=p(a)$ where $p'<0$ and $p''>0$ for $a\geq 0$. Eventually, the probability would converge to $0$ as $a\to \infty$. Under these assumptions, we obtain the following theorem:

\begin{theorem}\label{theorem:hidden}
    There exist a finite positive optimal effort $0<a^\diamond<\infty$ if the cost and error probability functions satisfy:
    \begin{equation}
        \begin{cases}
            c'(a^\diamond)+p'(a^\diamond)\Bar{L}=0\\
            c(a^\diamond)<(1-p(a^\diamond))\Bar{L}
        \end{cases}
        \label{eq:optcondi}
    \end{equation}
    where $\Bar{L}=(1-\rho^*)D_{AI}(l_{FP}+l_{FN})$. Otherwise, the optimal effort is $a^\diamond=0$. The proof can be found in Appendix~\ref{ap:optcondi}.
\end{theorem}
The theorem suggests that the cost and error probability function should be reasonable enough such that the company is willing to put effort to reduce the risk instead of completely counting on the insurer to cover the risks.

\section{AI Insurability Analysis}
\label{sec:result}

To establish AI liability insurance, we first need to check whether the AI-inflicted risks fulfill the insurability criteria. Berliner \cite{berliner1985large} proposed nine insurability criteria under three categories. We analyze each requisite and list down the criteria that might be problematic, as shown in Table~\ref{table:insurability}. In the following section, we will discuss the actuarial and market insurability requirement in our model. The social aspects of AI insurance are beyond the discussion of this work.

\renewcommand{\arraystretch}{1.4}{
\begin{table}[!th]
\footnotesize
\caption{Insurability criteria assessment for AI-inflicted risks in E-diagnosis systems.}
\begin{tabular}{||l|l|p{3.35cm}|p{3.3cm}|l|}
\hline
\rowcolor{Gray}
\multicolumn{1}{||c|}{\textbf{Type}} & \multicolumn{1}{c|}{\textbf{Criterion}} & \multicolumn{1}{c|}{\textbf{Characteristic}} & \multicolumn{1}{c|}{\textbf{Assessment}} &  \\ \hline
 & (1) Loss occurrence                                                 & Independent                                                          & Yes                                                              &  \\ \cline{2-4}
 & (2) Maximum possible loss                                            & Manageable                                                           & Quality requirement                                               &  \\ \cline{2-4}
& (3) Average loss per event                                          & Moderate                                                             & Quality requirement                                                          &  \\ \cline{2-4}
& (4) Loss exposure                                                   & Large enough                                                         & Yes                                                              &  \\ \cline{2-4}
\multirow{-5}{*}{Actuarial}                                     & (5) Information asymmetry                                           & Not excessive                                                  & Need regulation                                                  &  \\ \cline{1-4}
 & (6) Insurance premium                                               & Affordable for insureds                                              & Quality requirement/ Premium adjustment                                                 &  \\ \cline{2-4}
\multirow{-2}{*}{Market}                                        & (7) Coverage limits                                                 & Acceptable for insureds                                              & Yes                                                  &  \\ \cline{1-4}
& (8) Public policy                                                   & Consistent with social values                                        & Inconclusive                                                     &  \\ \cline{2-4}
\multirow{-2}{*}{Society}                                       & (9) Legal restrictions                                              & Not violated                                                         & Inconclusive                                                     &  \\ \cline{1-4}
\end{tabular}
\label{table:insurability}
\end{table}}

\subsection{Quality requirement}

One major problem that might put the insurability of AI-powered E-diagnosis systems in jeopardy is related to the quality of the system. In our framework, AI insurance for e-diagnosis systems might be uninsurable due to its classification errors. If the classification system performs worse than random guessing, it is unlikely for the insurance company to participate. The maximum possible loss (2) and average loss (3) would be very large as the loss is very likely to occur. According to U.S. Food and Drug Administration (FDA), the at-home COVID-19 test kit is required to have at least $80\%$ accuracy. The AI E-diagnosis system that fails to satisfy the requirement should not be insured or even be released to the public in the first place. 

Besides, the insurance company also needs to consider whether the premium value is affordable for the agent. Let the maximum acceptable premium of the market be $\Bar{\pi}$. According to Theorem~\ref{theorem:full}, the premium value is bounded by the expected loss of the system. If the AI system is at high risk of the wrong diagnosis, the expected loss $\mathbb{E}[x(a)]$ would increase and result in a higher premium range. If the lower bound of the premium range is larger than the maximum acceptable premium in the market, i.e., $\rho\E[x(a)]>\Bar{\pi}$, the insurance would not be established.

To control the potential loss of the AI system, the insurance company can incentives the insureds to minimize risky behaviors by setting prerequisite standards for the design of an AI entity. For instance, the insurance company can require the AI company to provide a quality control inspection report to
keep track of errors and ensure the quality of the product. This step ensures the company provides the product with promised quality.

Consider full information insurance benchmark (effort $a$ is known) with full coverage ($\rho=1$$)$. Based on the discussion above, we propose the following:
\begin{definition}[AI-powered E-diagnosis Insurability Requirement]
\label{def:require}
The AI-powered e-diagnosis system is insurable if it satisfies the following:
\begin{enumerate}
    \item The classification accuracy (ACC) is above FDA requirement:
    \begin{align*}
        ACC = \frac{p_t+1-p_f}{2} > 0.8
    \end{align*}
    \item The premium value is below the maximum acceptable market price for the agent, i.e., $\mathbb{E}[ x(a)]\leq \overline{\pi}$.
    \item The AI system should satisfy all quality prerequisites in order to get insured. 
\end{enumerate}
\end{definition}

\begin{figure}[t!]
\centering     
\subfloat[High Uncertainty]{\includegraphics[width=0.5\textwidth]{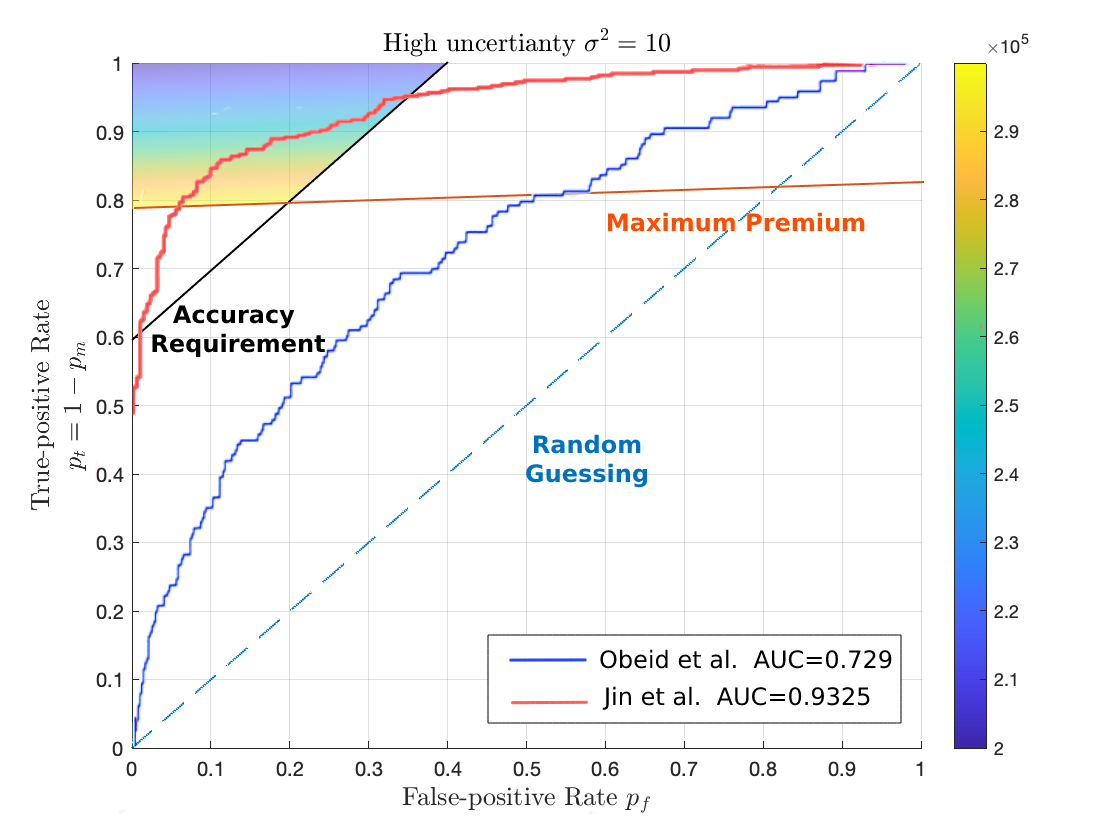}}
\subfloat[Low Uncertainty]{\includegraphics[width=0.5\textwidth]{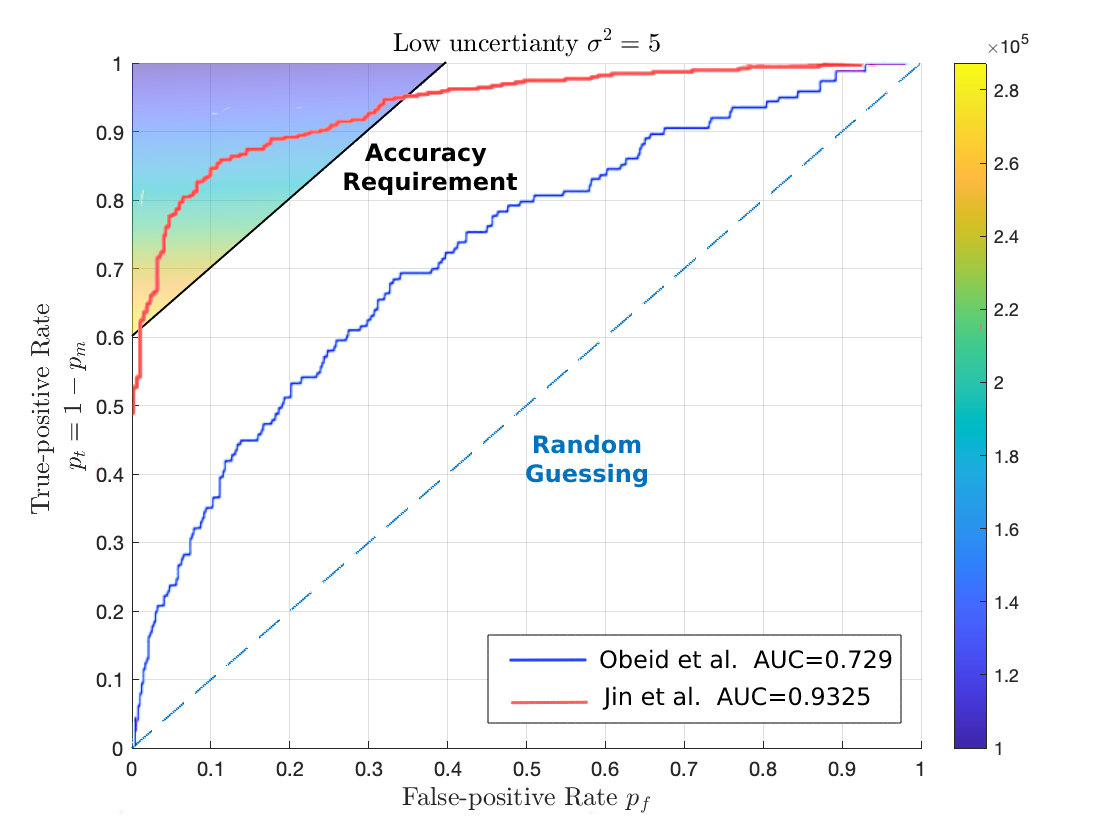}}
\caption{AI-powered E-diagnosis insurability requirement under different uncertainties. }
\label{fig:insurability}
\end{figure}

\noindent\textbf{Liability insurance serves as certification of AI}

In our framework, the quality of the AI-powered E-diagnosis system can be illustrated in the ROC plot. Figure~\ref{fig:insurability} illustrates the insurable region of the E-diagnosis system under different inherent uncertainties. We let the distribution size $D_{AI}=100$, loss value due to uncertainty $k=200$, and the maximum acceptable premium price $\Bar{\pi}=\$300,000$ to illustrate the results. 

As illustrated in Figure~\ref{fig:insurability}, when the AI system has high inherent uncertainty $\sigma^2=10$, both accuracy constraint and maximum premium constraint are activated. To compensate for the large uncontrollable loss caused by high inherent uncertainty, the AI system is required to generate more accurate testing results to reduce performance error. 
When the inherent uncertainty is low $\sigma^2=5$, the feasible ROC region is enlarged and only the accuracy constraint is activated. It is notable to mention that with low uncertainty, the premium value is also decreasing for the AI system with the same quality $(p_t,p_f)$ on the ROC curve.

We use two AI-powered COVID-19 diagnosis systems to illustrate the relationship between insurability requirements and AI quality \cite{obeid2020artificial,jin2020development}. We use the test data ROC curve in \cite{obeid2020artificial} and the ROC on MosMedData cohort \cite{morozov2020mosmeddata} in \cite{jin2020development}. The diagnosis system developed by Jin et al. possesses higher quality than the other system as the AUC is higher and closer to $1$. From the figure, Obeid's model is not insurable in both cases as it fails to satisfy the FDA accuracy requirement. Jin's model is able to get insured in both cases but the feasible threshold region is smaller when the inherent AI uncertainty is higher. 

By considering the inherent uncertainty and the insurability requirements, we ensure the loss of the AI system is under control. In this sense, AI liability insurance serves as a certification role of AI services. If the AI product is insured, the quality of the product is guaranteed. This encourages AI developers to improve system reliability, create low-risk algorithms, and provide better services. Well-designed AI liability insurance has the potential to mitigate AI risks and facilitate wide the adoption of such technologies.  Establishing AI liability insurance helps build a virtuous circle that encourages AI innovation and development, stimulates AI adoption, and eventually encourages the adoption of the liability insurance market. 

\vskip 3mm
\noindent\textbf{Quality regulation through liability insurance}

As suggested in the third point in Definition~\ref{def:require}, the insurance company can pose a set of quality prerequisites for the AI company to get insured. By including the requirements in the insurance contract, the insurer can inform the AI company of suggestions to reduce AI-inflicted losses. In this way, liability insurance can act as quasi-regulators and exert a behavioral channeling effect on the insureds \cite{baker2012regulation}. The regulatory requirements ensure the performance of AI products and incentives the AI company to minimize risky behaviors. This, in turn, will help reduce the cost of insurance policies and assist the development of the AI insurance market. It is noteworthy that the insurance company needs to have expert knowledge of AI technologies to provide effective and professional advice.

\subsection{Premium adjustment}

Emerging technologies like AI have always been a challenge for the insurance market as the associated risks are usually high at the early stage. It creates difficulty in the insurability of AI systems as the initial premium may be unaffordable for most companies. Fortunately, the risks related to inherent uncertainty can be reduced with time when people have more knowledge about AI. This requires the insurance company quickly identify the current uncertainty level and flexibly adjust the premium based on actuarial data. We analyze the premium adjustment in AI liability insurance and provide suggestions on it.

\begin{figure}[!ht]
    \centering
    \includegraphics[width=0.7\linewidth]{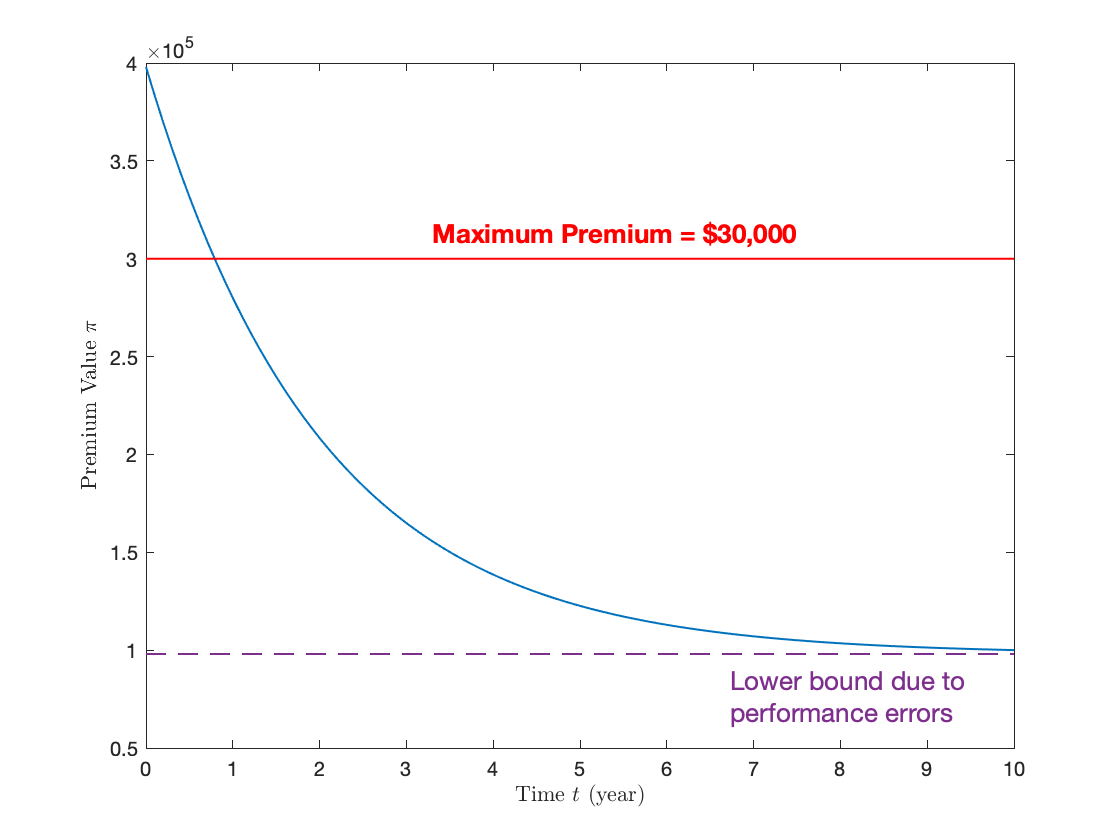}
    \caption{Premium adjustment. As the inherent uncertainty in AI technologies decreases with time, the insurance premium value should adjust dynamically. }
    \label{fig:adjust}
\end{figure}

We consider a full information insurance model where $a$ is given and system error rates are $p_t=0.9$ and $p_f=0.1$. We assume the evolution of uncertainty $\theta$ follows \eqref{eq:theta} and $\sigma^2=15$ and $m=0.5$. Figure~\ref{fig:adjust} illustrate the risk-adjusted premium values at different times (the unit on the x-axis is the year). 

At the early stage, the uncertainty is high ($\sigma^2(t)$ is large), leading to a high uncertainty risk $l_\theta$ and premium. The early adopters of AI technology need to pay high premiums that might exceed the market average. This is the price they pay for their curiosity and willingness to take more risks. This indicates that the early pursuers of AI insurance would be risk-loving agents. Later on, the uncertainty of AI will decrease. The insurance company can adjust the premium so that the risk-averse agents can also afford the premium and join the risk pool. In our model, the inherent uncertainty is independent of the performance risks $l_{AI}$. Thus, there exists a lower bound on the premium adjustment due to the classification performance error.

\vskip 3mm
\noindent\textbf{The need of government support}

The insurance premium requirement in the insurability criteria will be satisfied as the inherent uncertainty level drops. However, the insurance market would need support from the government at the initial stage. If the profitability of the insurance company is not achievable (the initial premium is too high that no agent wants to participate), the government can offer subsidized premiums to bridge the gap at the beginning \cite{lior2022insuring}. The government support would allow insurers to offer reasonable policies and still make a profit. In this way, AI liability insurance can be established even with high inherent AI uncertainty.

\subsection{Moral Hazard Problem}

Finally, we discuss the moral hazard problem in AI liability insurance due to information asymmetry. The problem of moral hazard arises as the agent has hidden action $a$ that is not observable to the insurer.  Suppose the company have two available effort choices $a\in\mathcal{A}=\{a_H,a_L\}$, where $a_H>a_L\geq 0$. Once we determine the insurance plan $(\pi^*,\rho^*)$, the first-best action is obtained by solving 
\begin{align}
    a^* \in \arg\max_{a\in\mathcal{A}} \int V\left(w_I + \pi^* -\rho^* x(a) \right) d F(x|a) 
\end{align}
Since the risk will be reduced when the company puts more effort, the insurer would prefer $a^*=a_H$ as the first-best action. 

For the agent, if we consider a risk-neutral agent with optimal diagnosis threshold in \eqref{eq:pfpm}, finding $a^\diamond$ is equivalent to solving:
\begin{align}
    \min_{a\in \{a_H,a_L\}} \quad c(a)+(1-\rho^*)D_{AI}Q\left( D h(a)\right)(l_{FP}+ l_{FN}).
\end{align}
Since $c'(a)>0$ and $d/da (Q(Dh(a)))<0$, we have the following proposition:

\begin{proposition}The second-best action will coincide with the first-best action when:
\begin{align}
    Q(Dh(a_L))-Q(Dh(a_H))\geq \frac{1}{\Bar{L}}\left[c(a_H)-c(a_L)\right],
\end{align}
where $\Bar{L}=(1-\rho^*)D_{AI}(l_{FP}+l_{FN})$. If we denote the change in misclassification probability as $\Delta (Pr_L-Pr_H)$ and the change in cost as $\Delta (c_H-c_L) $, we have
\begin{align}
    \Delta (Pr_L-Pr_H)\geq \frac{1}{\Bar{L}}\Delta (c_H-c_L).
\end{align}
\label{prop:mh}
\end{proposition}

\noindent\textbf{Suggestions to reduce moral hazard}

The proposition implies that the company would prefer to spend more effort $a = a_H$  when it largely reduces the error probability while not bringing too much cost. It suggests that efficient effort in reducing risks and reasonable cost of the effort would help reduce the moral hazard problem in the AI liability insurance market. The insurance company as a regulation mechanism can provide loss prevention advice to the AI company on how to modify their actions in order to efficiently improve system performance and avoid losses. Our discussion in this work focuses on the linear insurance contract. In practice, besides the regulation via insurance, the insurer can also mitigate the risks of moral hazard through insurance contract design. For instance, deductibles, coinsurance, coverage limitations, etc.

\section{Conclusion}
\label{sec:conclusion}

In this work, we analyze AI liability insurance with an example in an AI-powered E-diagnosis system. We provide a quantitative risk assessment model with evidence-based numerical analysis. We discuss the insurability criteria of AI-powered technologies and provide suggestions based on the analysis. We suggest building upon existing insurance frameworks with necessary adjustments to accommodate the unique features of AI products. By including quality requirements in the insurance contract, AI liability insurance can act as a regulatory mechanism to incentivize compliant behaviors of AI entities. Liability insurance also serves as a certificate of AI quality and motivates developers to create high-quality AI products going forward. Furthermore, we suggest a dynamic premium adjustment to reflect the dynamic evolution of the inherent uncertainty in AI technologies. Support from the government is necessary to ensure the profitability of insurance companies. Moral hazard problems are discussed and suggestions for AI liability insurance are provided.  In general, well-designed AI liability insurance products will play a pivotal role in the innovation and development of the AI ecosystem and help the adoption of AI in various areas. 

For future work, we would investigate AI insurance under adversarial attacks and discuss the intersections and differences from cyber insurance.  Accountability identification in the AI supply chain is another interesting topic to consider.

\appendix
\section{Proof of Theorem 1}
\label{ap:pro}
\begin{proof}

To ensure the profitability of the insurer, we should have
\begin{align}
 \max_{\pi,\rho,a} \quad & \int V\left(w_I + \pi -\rho x(a) \right) d F(x|a) \geq V(w_I).
\end{align}
For a risk-neutral insurer, the constraint can be reduced to
\begin{align}
    w_I+\pi-\rho\E[x(a)]\geq w_I \quad \Rightarrow \quad \rho\E[ x(a)]\leq \pi,
\end{align}
which gives the left inequality in Theorem~\ref{theorem:full}.

Then, consider the (IR) constraint with $\underline{U}=\E[U(w_A-c(a)-x(a))]$,
\begin{align}
     \int U\left(w_A - c(a)-\pi-(1-\rho)x\right) d F(x|a) \geq \underline{U}.
\end{align}
Apply the exponential utility function with $u(x) = 1- exp(-\epsilon x)$ into both sides, we obtain
\begin{align}
    \mathbb{E}[e^{-\epsilon(w_A-x(a))}] \geq \mathbb{E}[e^{-\epsilon(w_A- \pi - (1-\rho)x(a))}] .
\end{align}
Further simplify the inequality, we have
\begin{align}
   \mathbb{E}[e^{\epsilon\pi}] \leq \mathbb{E}[e^{\epsilon\rho x(a))}].
\end{align}
Since the expectation is only related to $x(a)$, taking the logarithm on both sides yields the right inequality in Theorem~\ref{theorem:full}:
\begin{align*}
    \pi \leq  \frac{1}{\epsilon}\log \mathbb{E}[e^{\epsilon\cdot\rho x(a)}].
\end{align*}
\end{proof}

\section{Proof of Theorem 2}
\label{ap:optcondi}
The objective function is 
\begin{align}
     f_V(a) =  c(a)+ p(a)\Bar{L},
\end{align}
where $\Bar{L}=(1-\rho^*)D_{AI}(l_{FP}+l_{FN})$. 

Under our assumptions, the objective function is a combination of convex functions $c(a)$ and $p(a)$, thus is still convex. For a convex function $f_V(a)$, if there exists a minimum point on the feasible region $a\in[0,+\infty)$, this point is a global minimum point. If $a^\diamond$ is the global minimum point, the first order condition requires:
\begin{align}
    c'(a^\diamond)+p'(a^\diamond)\Bar{L}=0,
\end{align}
which is the first requirement in the theorem. 

Furthermore, we hope that the global minimum point is between the interval of the extreme points $0$ and $\infty$. When the company put no effort into development, i.e., $a=0$, the cost should be $0$ and the error probability should be $1$. Thus, the value of the objective function is
\begin{align}
    f_V(0)=c(0)+p(0)\Bar{L} = \Bar{L}.
\end{align}
When the company put an extremely large effort into development, i.e., $a\to+\infty$, the cost should be close to $+\infty$ and the error probability should approach $0$. Thus, the value of the objective function is
\begin{align}
    \lim_{a\to+\infty} f_V(a)=c(a)+p(a)\Bar{L} = +\infty.
\end{align}
To ensure the global optimum satisfies $0<a^\diamond<\infty$, we need
\begin{align*}
    f_V(a^\diamond)=c(a^\diamond)+p(a^\diamond)\Bar{L} < \Bar{L},
\end{align*}
as $\Bar{L}<+\infty$. This is the second requirement in the theorem. With these two requirements, we can prove the existence of a finite positive optimal effort as $0<a^\diamond<\infty$.


\bibliographystyle{splncs04}
\bibliography{ref}

\end{document}